\documentclass[10pt,twocolumn,letterpaper]{article}

\usepackage{cvpr}
\usepackage{times}
\usepackage{epsfig}
\usepackage{graphicx}
\usepackage{amsmath}
\usepackage{amssymb}
\usepackage{subfig}
\usepackage{graphicx}
\usepackage{booktabs}
\usepackage{array}
\usepackage{rotating}
\usepackage{makecell}
\usepackage{wrapfig}
\usepackage{framed}
\usepackage{enumitem}
\usepackage{balance}
\usepackage{tabularx}
\usepackage{ragged2e}
\usepackage{afterpage}

\usepackage[loop,autoplay]{animate} %
\newif\ifcompileanimated
\compileanimatedtrue  %

\usepackage{adjustbox} %
\usepackage{enumitem} %

\usepackage[pagebackref=true,breaklinks=true,colorlinks,bookmarks=false]{hyperref}

\hypersetup{
	pdftitle={CED: Color Event Camera Dataset},
	pdfsubject={Computer Vision, Event-based Vision},
	pdfauthor={Cedric Scheerlinck, Henri Rebecq, Timo Stoffregen, Nick Barnes, Robert Mahony, Davide Scaramuzza}
	pdfkeywords={Event cameras, Dynamic Vision Sensor, Color, Dataset, Image reconstruction}
}

\cvprfinalcopy %

\def\cvprPaperID{20} %

\usepackage{cite}
\usepackage[per-mode=symbol]{siunitx}
\DeclareSIUnit\revpermin{rpm}

\newcommand\nomarkerfootnote[1]{%
  \begingroup
  \renewcommand\thefootnote{}\footnote{#1}%
  \addtocounter{footnote}{-1}%
  \endgroup
}

\usepackage{amsfonts} %

\global\long\def\pixLocation{(x,y)} %
\global\long\def\colorKeyboardLong{Color Keyboard} %
\global\long\def\flyingRoom{Flying Room} %
\global\long\def\jengaUnbroken{Jenga} %
\newcommand*{\theadCustom}[1]{\multicolumn{1}{c}{\textbf{#1}}}
\newcolumntype{R}[1]{>{\RaggedLeft\arraybackslash}p{#1}}
\newcolumntype{C}[1]{>{\centering\arraybackslash}p{#1}}

\usepackage{algorithm}
\usepackage[noend]{algpseudocode}

\usepackage{xcolor}

\definecolor{somegray}{rgb}{0.5, 0.5, 0.5}
\newcommand{\darkgrayed}[1]{\textcolor{somegray}{#1}}
\makeatletter
\newcommand*\titleheader[1]{\gdef\@titleheader{#1}}
\AtBeginDocument{%
	\let\st@red@title\@title
	\def\@title{%
		\vskip-4.5em
		\bgroup\normalfont\normalsize\centering\@titleheader\par\egroup
		\vskip2.8em\st@red@title}
}
\makeatother

\titleheader{\darkgrayed{This paper has been accepted for publication at the IEEE Conference on Computer Vision \\
		and Pattern Recognition Workshops (CVPRW), Long Beach, 2019.
		\copyright IEEE}}

\title{CED: Color Event Camera Dataset}

\author{
Cedric Scheerlinck \textsuperscript{\dag *}
\and
Henri Rebecq \textsuperscript{\ddag *}
\and
Timo Stoffregen \textsuperscript{\S}
\and
Nick Barnes \textsuperscript{\dag}
\and
Robert Mahony \textsuperscript{\dag}
\and
Davide Scaramuzza \textsuperscript{\ddag}
}
\usepackage[normalem]{ulem} %

\begin{document}

\maketitle

\nomarkerfootnote{\textsuperscript{*} Equal contribution.}
\nomarkerfootnote{\textsuperscript{\dag} Australian National University, Canberra, ACT, Australia.}
\nomarkerfootnote{\textsuperscript{\ddag} Dept.~Informatics, Univ.~of Zurich and Dept.~Neuroinformatics, Univ.~of Zurich and ETH Zurich.}
\nomarkerfootnote{\textsuperscript{\S} Monash University, Melbourne, VIC, Australia.}

\begin{abstract}
Event cameras are novel, bio-inspired visual sensors, whose pixels output asynchronous and independent timestamped spikes at local intensity changes, called `events'.
Event cameras offer advantages over conventional frame-based cameras in terms of latency, high dynamic range (HDR) and temporal resolution.
Until recently, event cameras have been limited to outputting events in the intensity channel, however, recent advances have resulted in the development of color event cameras, such as the Color-DAVIS346.
In this work, we present and release the first \emph{Color Event Camera Dataset (CED)}, containing 50 minutes of footage with both color frames and events.
CED features a wide variety of indoor and outdoor scenes, which we hope will help drive forward event-based vision research.
We also present an extension of the event camera simulator ESIM \cite{Rebecq18corl} that enables simulation of color events.
Finally, we present an evaluation of three state-of-the-art image reconstruction methods that can be used to convert the Color-DAVIS346 into a continuous-time, HDR, color video camera to visualise the event stream, and for use in downstream vision applications.
\end{abstract}

\noindent 
\textbf{Website:} \url{http://rpg.ifi.uzh.ch/CED}

\section{Introduction}
\label{sec:intro}

\global\long\def\frontWidth{4.05cm} %
\begin{figure}[t]
	\centering
	\begin{tabular}{c c}
    \includegraphics[width=\frontWidth]{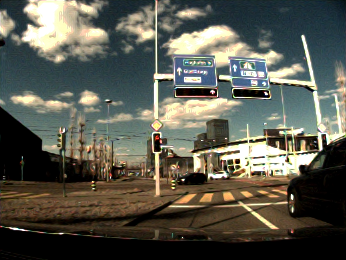}
    \hspace{0.1pt}
    & \includegraphics[width=\frontWidth]{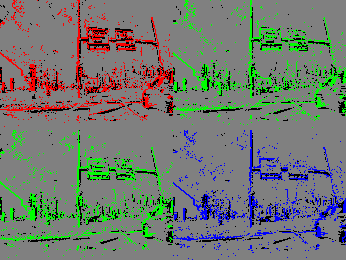}\\
    
   	\includegraphics[width=\frontWidth]{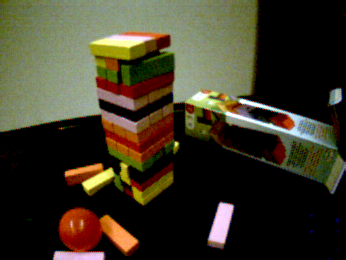}
   	\hspace{0.1pt}
   	& \includegraphics[width=\frontWidth]{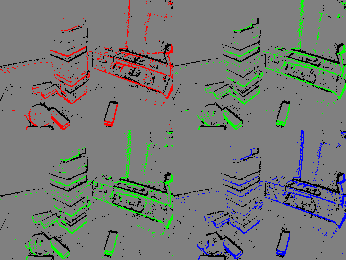}\\
    	
    \end{tabular}
\caption{Our Color Event Camera Dataset (CED) features both outdoor (top row) and indoor (bottom row) sequences, and provides color images (left column) and color events (right row) from the Color-DAVIS346 for each sequence.
}
\label{fig:frontpage}
\end{figure}%

Since their recent addition to the computer vision community \cite{Lichtsteiner08ssc}, event cameras have challenged conventional thinking about how to solve computer vision problems.
Instead of producing global-shutter images at a fixed frame-rate as in conventional cameras, event cameras have pixels that operate independently and asynchronously.
When the brightness change at a given pixel exceeds a threshold, that pixel emits an event containing its $\pixLocation$ address, timestamp and polarity.
Event cameras offer several advantages; 
they sample at the rate of scene dynamics without having to wait for an external shutter cycle, and the output is data-driven and non-redundant.
This means that event cameras have extremely low latency, low power consumption and bandwidth requirements, high dynamic range and suffer essentially no motion blur.
The temporal resolution of current event cameras is in the order of microseconds.

Since their introduction, event cameras have spawned a flurry of research. They have been used in 
feature detection and tracking
\cite{
	Gehrig18eccv, 
	Barranco18iros,  
	Alzugaray18ral, 
	Scheerlinck19ral
}, 
depth estimation \cite{
	Rebecq16bmvc, 
	Rebecq17ijcv, 
	Kim16eccv, 
	Gallego18cvpr
}, 
stereo \cite{
	Zou17bmvc, 
	Andreopoulos18cvpr, 
	Zhu18eccv, 
	Zhou18eccv
}, 
optical flow \cite{
	Benosman14tnnls, 
	Mueggler15icra, 
	Stoffregen17acra, 
	Liu18bmvc
}, 
image reconstruction \cite{
	Kim14bmvc, 
	Barua16wcav, 
	Bardow16cvpr, 
	Reinbacher16bmvc, 
	Munda18ijcv, 
	Scheerlinck18accv,
	Pan19cvpr
}, 
localization \cite{
	Gallego17pami, 
	Reinbacher17iccp, 
	Mueggler17ijrr, 
	Bryner19icra
},
SLAM \cite{
	Kueng16iros, 
	Rebecq17ral, 
	Gallego17ral
}, 
visual-inertial odometry \cite{
	Mueggler17tro, 
	Zhu17cvpr, 
	Rebecq17bmvc, 
	Rosinol18ral
}, 
pattern recognition \cite{
	Sironi18cvpr, 
	Maqueda18cvpr, 
	Zhu18rss, 
	Shrestha18nips
}, 
and more.
In response to the growing needs of the community, several important event-based vision datasets have been released, directed at popular topics such as SLAM \cite{Mueggler17ijrr}, optical flow \cite{Zhu18ral,Rueckauer16fns} and recognition \cite{Orchard15fns,Sironi18cvpr}.
Event camera datasets enable better benchmarking and reproducibility, and grant researchers access to high quality event data in a range of environments without necessarily having to acquire an expensive event camera.
\setlength{\intextsep}{0pt}%
\setlength{\columnsep}{14pt}%
\begin{wrapfigure}{r}{0.2\textwidth}
	\includegraphics[trim={20cm 0 10cm 10cm},clip,width=0.2\textwidth]{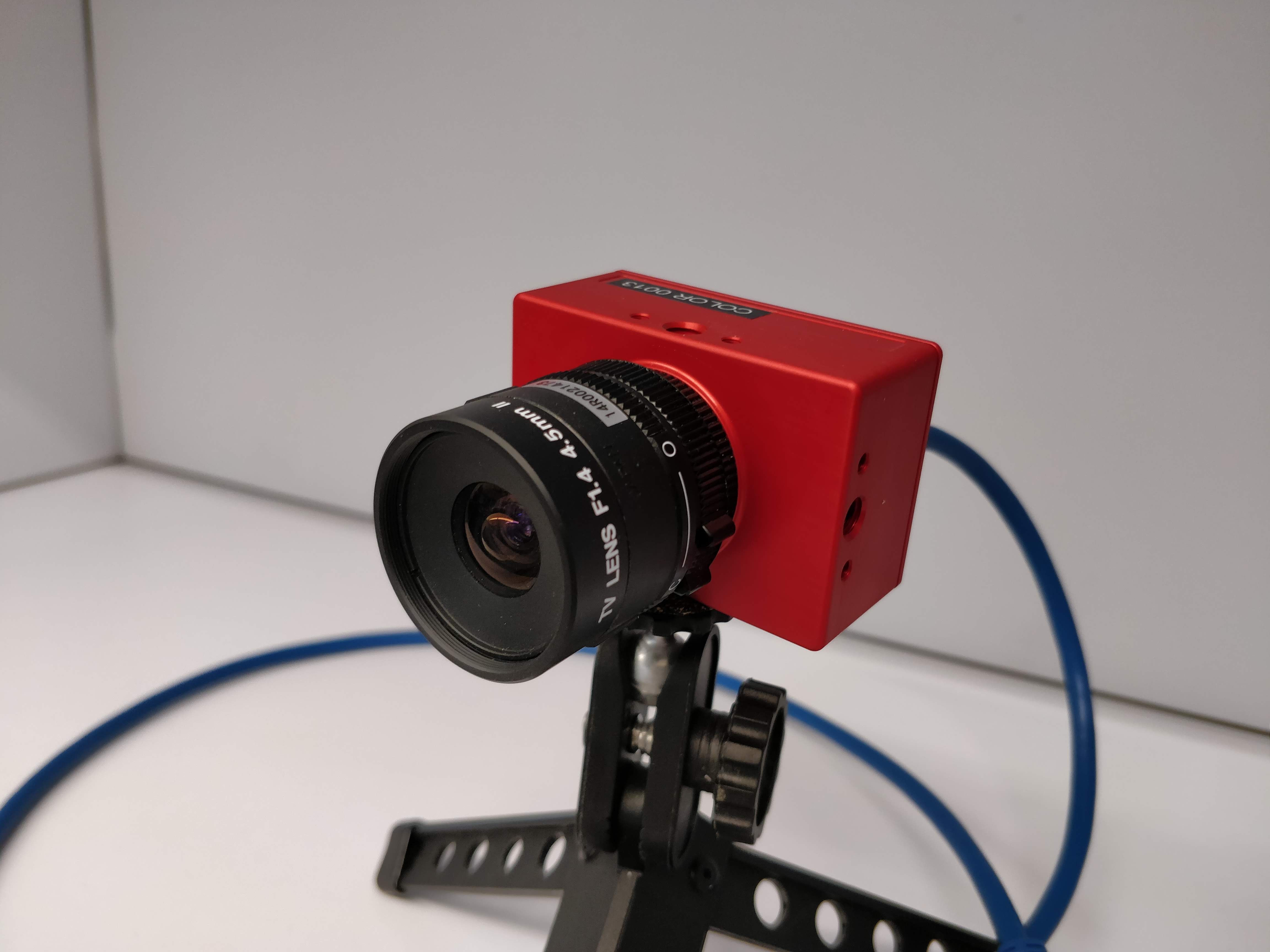}
	\caption{``DAVIS346 Red Color" camera used for dataset collection.}
	\label{fig:davis}
	\vspace{0.4cm}%
\end{wrapfigure}%
While existing datasets are limited to monochrome events, event camera technology has since advanced to allow color events and frames \cite{Taverni18tcsii}, which opens the door to a new generation of color event processing.

The addition of color information to event-based vision has the potential to improve performance of many tasks, such as segmentation \cite{Marcireau18fns} and recognition, where it is known that color is an important source of visual information \cite{Tremeau08jivp}.
Early works have shown promising results using prototype color event cameras \cite{Moeys17iscas,Moeys17tbcas,Li15iscas}, or a mirrored-rig with three monochrome cameras and three color filters \cite{Marcireau18fns}, however, to-date there are no publicly available color event datasets.
Further, the wider research community has limited access to color event cameras, hindering progress into color event vision research.

We present the first \emph{Color Event Camera Dataset} (Fig.~\ref{fig:frontpage}) that aims to spur research into color event vision by providing the community with high quality color event data, alongside color frames from the Color-DAVIS346 \cite{Taverni18tcsii}.
The Color-DAVIS346 (Fig.~\ref{fig:davis}) is the latest color event camera, built upon the popular line of DAVIS cameras that many existing datasets and research is based off.
Rather than directing our focus at a specific target application, we aim to cater for general purpose vision research by including a diverse range of scenes (simple objects, indoor/outdoor scenes, people), lighting conditions (daylight, indoor light, low-light), camera motions (linear, 6-DOF motion) and dynamics.
While we do not provide ground truth labels for any specific task (\eg optical flow estimation, object detection, etc.), we provide color images from the sensor that are naturally synchronized and registered to events.
These images may be used to generate proxy labels for any task of interest (using either conventional computer vision, or manual annotation) that can be transferred to the events.

To visually unveil the color information contained in color events, we evaluate and compare three state-of-the-art event-based image reconstruction methods \cite{Reinbacher16bmvc,Scheerlinck18accv,Rebecq19cvpr} on our Color Event Camera Dataset.
Image reconstruction is an active field of event-based vision research
\cite{
	Kim14bmvc,
	Barua16wcav,
	Bardow16cvpr,
	Reinbacher16bmvc,
	Munda18ijcv,
	Scheerlinck18accv,
	Scheerlinck19ral,
	Rebecq19cvpr
}
that allows visualisation of the event stream, and enables application of decades of computer vision research and expertise on event data, which in its raw form is inaccessible to powerful tools such as convolutional neural networks.
Further, event reconstructed images have the potential to retain desirable qualities of event cameras, such as high dynamic range, high temporal resolution and immunity to motion blur.

\textbf{Contributions:}
\begin{enumerate}[topsep=3pt,itemsep=0pt]
	\item We present CED: Color Event Camera Dataset containing 50 minutes of both color events and frames in a wide range of natural scenes with static and dynamic objects, and covering a variety of camera-motions from simple translations and rotations to unconstrained 6-DOF motions.
	\item We release a color event camera simulator, based on ESIM \cite{Rebecq18corl}.
	\item We present color video reconstructions from a color event camera, comparing three state-of-the-art reconstruction methods.
	Video reconstruction provides a natural way to visualize the event stream and enable image-based processing on events.
\end{enumerate}
\section{Related Works}
\label{sec:related}

Many event-based vision datasets have been published since the introduction of the DVS \cite{Lichtsteiner08ssc}.
Most of these datasets were recorded using a DAVIS \cite{Brandli14ssc} event camera or similar and have a particular use-case in mind, such as image reconstruction \cite{Scheerlinck18accv}, recognition \cite{
	Orchard15fns,
	Sironi18cvpr,
	Serrano-Gotarredona15fns
},
optical flow \cite{
Rueckauer16fns,
Mitrokhin18iros,
Bardow16cvpr
},
driving/SLAM \cite{
	Zhu18ral,
	Bryner19icra,
	Gallego17pami
}.
The dataset perhaps most similar to ours is the \emph{Event-Camera Dataset and Simulator} \cite{Mueggler17ijrr}.
All of the above datasets are limited to monochrome temporal contrast or gray-level events.
Our Color Event Camera Dataset (CED) doesn't have a particular use-case in mind and aims simply to cover a wide range of scenarios and motions that can be used in a broad swathe of research topics.

The need for publicly available datasets of arbitrary event data is partly driven by the fact that event cameras are scarce and expensive hardware acquisitions.
For this reason several event camera simulators have been developed in previous years, the most sophisticated of which is the \emph{ESIM} \cite{Rebecq18corl}.
While ESIM provides high quality, realistic event data and ground-truth from a free moving simulated camera in an arbitrary 3D modeled environment, it does not support color events.
Nor does (to our knowledge) any other contemporary, publicly available event simulator.
We propose an extension of ESIM to simulate color events and make it publicly available.

Thus far there have been few works that use color events. 
One particular counterexample is Marcireau \etal \cite{Marcireau18fns}, who perform color segmentation on color events.
However, in this work the authors felt compelled to build their own color event camera using a complex array of beam splitting mirrors and filters to channel light into three separate event cameras.
Further, this setup did not allow capturing color frames, which had to be instead reconstructed from the event streams of the three sensors.
Our dataset hopes to save future researchers this kind of effort.

The C-DAVIS \cite{Li15iscas} was one of the first color event cameras, based on the DAVIS \cite{Brandli14ssc} with VGA resolution color (RGBW) frames and QVGA monochrome events.
The SDAVIS192 \cite{Moeys17tbcas} had improved sensitivity over the DAVIS, able to output color (RGBW) events and frames at \mbox{188 $\times$ 192} pixel resolution.
Moeys \etal \cite{Moeys17iscas} used the SDAVIS192 to demonstrate color image reconstruction from events using 1) na\"ive integration and 2) Poisson integration \cite{Agrawal05iccv} of a gradient field based on the surface of active events \cite{Benosman14tnnls}.
The Color-DAVIS346 \cite{Taverni18tcsii} is the latest color event camera at the time of writing, and outputs color (RGBG) events and frames at 346 $\times$ 260 resolution.

\section{CED: Color Event Camera Dataset}
\label{sec:dataset}

\global\long\def\bayerHeight{3.2cm} %
\begin{figure}[t]
	\centering
	\begin{tabular}{cc}
	\includegraphics[height=\bayerHeight]{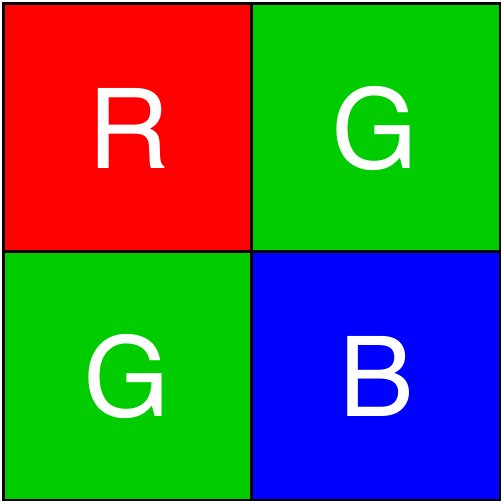}
	\hspace{1pt}
    & \includegraphics[height=\bayerHeight]{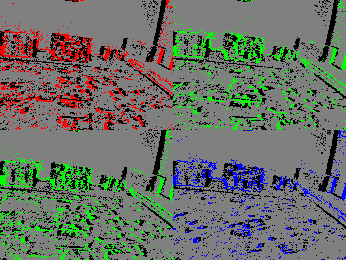}
	\end{tabular}
	\caption{Left: 2$\times$2 RGBG Bayer pattern in the Color-DAVIS346.
		Right: Events from the Color-DAVIS346 split into each color.
		Positive (ON) events are colored by the corresponding filter color, negative (OFF) events are black.
	}
	\label{fig:bayer}
\end{figure}%

The Color-DAVIS346 \cite{Taverni18tcsii} consists of an 8$\times$6mm CMOS chip patterned with RGBG filters (Fig.~\ref{fig:bayer}), able to output color events and standard frames at 346$\times$260 pixel resolution.
Table~\ref{tab:biases} displays the camera bias settings used (based off the defaults provided in the DAVIS ROS driver%
\footnote{\url{https://github.com/uzh-rpg/rpg_dvs_ros}}).
Events generated by the DAVIS are reported with microsecond timestamp precision.
We provide time-stamped, raw frames from the DAVIS, as well as color frames obtained via demosaicing \cite{opencvDemosaicing}.
To minimize motion blur in the DAVIS frames, we use fixed exposure fine-tuned for each indoor sequence.
We use auto-exposure for outdoor sequences since it is bright enough to drive exposure time down.
No infrared filter is used unless otherwise specified.
We provide binary (rosbag) files containing synchronized and time-stamped events, raw images and color images.

The Color Event Camera Dataset (Fig. \ref{fig:dataset_previews}) contains 50 minutes of footage consisting of 100k color DAVIS frames and over one billion color events.
The sequences cover a wide variety of scenes that showcase some of the key properties of the technology, namely high dynamic range, high temporal resolution and immunity to motion-blur.
We include five categories (Table \ref{tab:types}): Simple, Indoors, People, Driving and Calibration.
Simple contains sequences in favorable conditions, \ie well-lit, moderate camera motions, where the DAVIS frame is typically sharp and well-exposed.
Indoors contains challenging conditions such as low-light, fast camera motion, as well as natural indoor office scenes.
People consists of pre-determined actions such as sitting, waving, dancing with both static and dynamic camera.
Driving is filmed through the windshield of a car in sunny conditions and contains a range of environments including highways, tunnels, city and country.
Calibration shows a ColorChecker and density step target in various lighting conditions including fluorescent, low-light, outdoors, with and without an infrared filter.
\setlength{\tabcolsep}{0.2cm} %
\begin{table}[t]
\centering
\caption{Bias settings used for the Color-DAVIS346.}
\label{tab:biases}
	\begin{tabular}{ c c c c c }
		\textbf{Bias} & \multicolumn{2}{c}{\textbf{Indoors}} & \multicolumn{2}{c}{\textbf{Outdoors}}\\
		
		\midrule
		 & Coarse & Fine & Coarse & Fine\\
		
		\texttt{DiffBn} & 4 & 39 & 4 & 39 \\
		
		\texttt{OFFBn} & 4 & 0 & 4 & 0 \\
		
		\texttt{ONBn} & 6 & 200 & 6 & 200 \\
		
		\texttt{PrBp} & 2 & 58 & 3 & 0 \\
		
		\texttt{PrSFBp} & 1 & 33 & 1 & 33 \\
		
		\texttt{RefrBp} & 4 & 25 & 4 & 25 \\
		\hline
	\end{tabular}
\vspace{-5mm}
\end{table}
\setlength{\tabcolsep}{0.05cm} %

\global\long\def\heightplot{2.8cm} %
\global\long\def\widthplot{3.72615cm} %
\global\long\def\vTextWidth{0.4cm} %
\global\long\def\mColumnWidth{3.75cm} %
\global\long\def\gapWidth{0.1cm} %
\begin{figure*}[t]
	\centering
	\begin{tabular}{  %
		>{\centering\arraybackslash}m{\vTextWidth}
		>{\centering\arraybackslash}m{\mColumnWidth}
		>{\centering\arraybackslash}m{\mColumnWidth}
		>{\centering\arraybackslash}m{\gapWidth}
		>{\centering\arraybackslash}m{\vTextWidth}
		>{\centering\arraybackslash}m{\mColumnWidth}
		>{\centering\arraybackslash}m{\mColumnWidth}
		}

    & \multicolumn{2}{c}{ \textbf{Simple}}
    & &
    & \multicolumn{2}{c}{ \textbf{People}}
    \\
   
    \rotatebox{90}{\colorKeyboardLong}
    & \includegraphics[width=\widthplot,height=\heightplot]{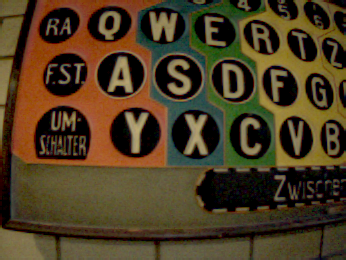}
    & \includegraphics[width=\widthplot,height=\heightplot]{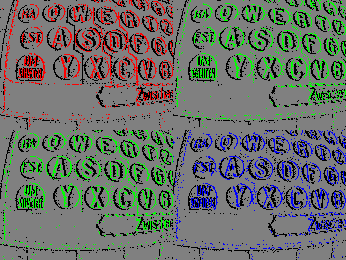}
    & & \rotatebox{90}{Air guitar}
    & \includegraphics[width=\widthplot,height=\heightplot]{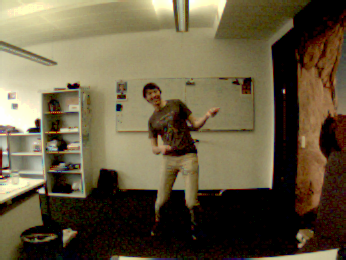}
    & \includegraphics[width=\widthplot,height=\heightplot]{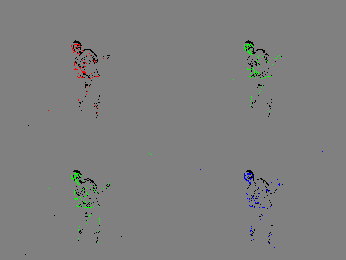}
    \\ 

    \rotatebox{90}{Fruits}
    & \includegraphics[width=\widthplot,height=\heightplot]{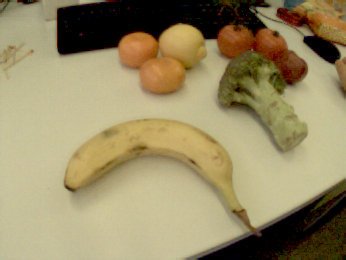}
    & \includegraphics[width=\widthplot,height=\heightplot]{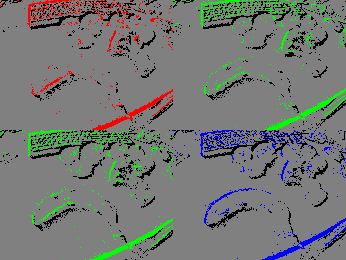}
    & & \rotatebox{90}{Jumping}
    & \includegraphics[width=\widthplot,height=\heightplot]{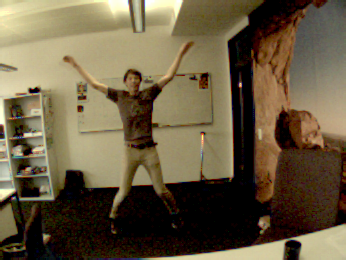}
    & \includegraphics[width=\widthplot,height=\heightplot]{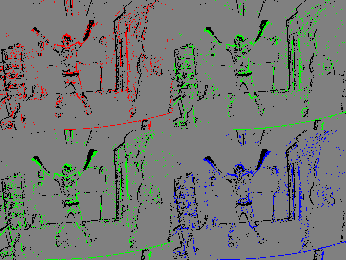}
    \\ 
    
    \rotatebox{90}{Wires}
    & \includegraphics[width=\widthplot,height=\heightplot]{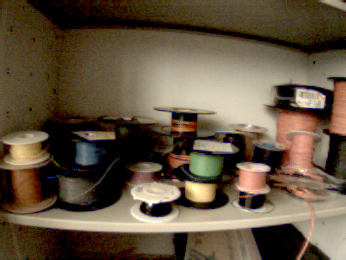}
    & \includegraphics[width=\widthplot,height=\heightplot]{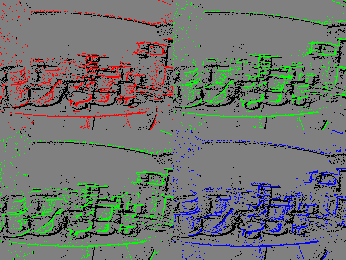}
    & & \rotatebox{90}{Selfie}
    & \includegraphics[width=\widthplot,height=\heightplot]{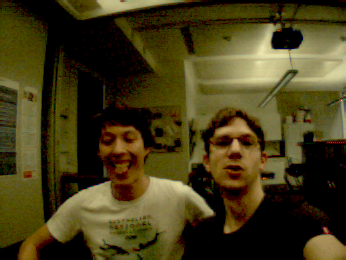}
    & \includegraphics[width=\widthplot,height=\heightplot]{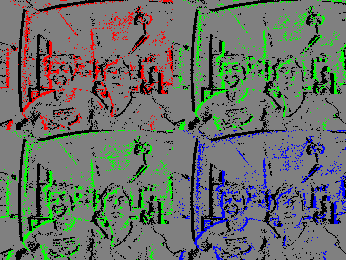}
    \\
	
    & \multicolumn{2}{c}{ \textbf{Indoors}}
    & &
    & \multicolumn{2}{c}{ \textbf{Driving}}
    \\
    
    \rotatebox{90}{Kitchen}
    & \includegraphics[width=\widthplot,height=\heightplot]{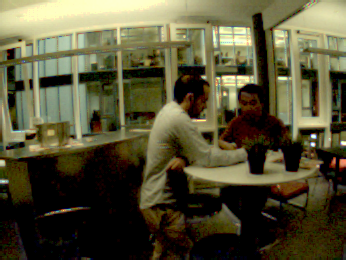}
    & \includegraphics[width=\widthplot,height=\heightplot]{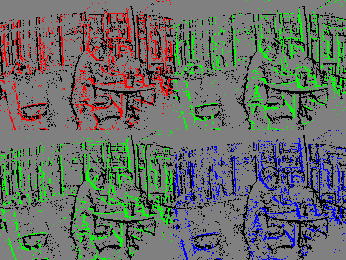}
    & & \rotatebox{90}{City 1}
    & \includegraphics[width=\widthplot,height=\heightplot]{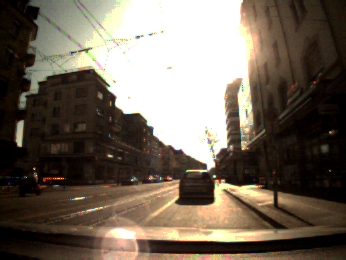}
    & \includegraphics[width=\widthplot,height=\heightplot]{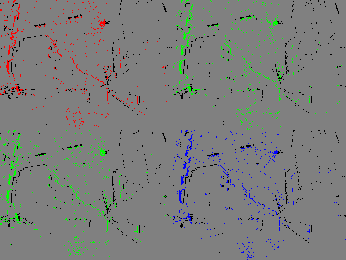}
    \\ 

    \rotatebox{90}{Office}
    & \includegraphics[width=\widthplot,height=\heightplot]{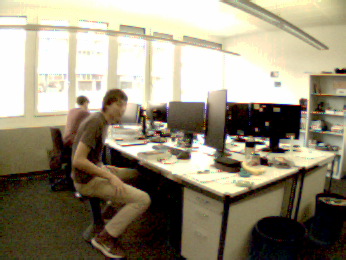}
    & \includegraphics[width=\widthplot,height=\heightplot]{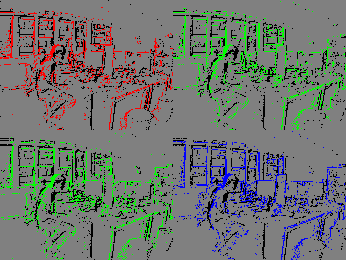}
    & & \rotatebox{90}{City 2}
    & \includegraphics[width=\widthplot,height=\heightplot]{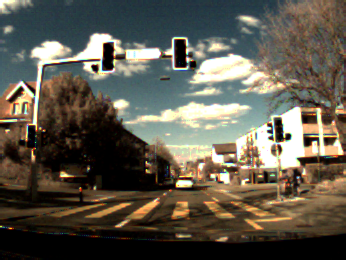}
    & \includegraphics[width=\widthplot,height=\heightplot]{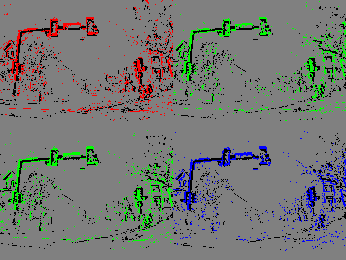}
    \\ 

    \rotatebox{90}{Foosball}
    & \includegraphics[width=\widthplot,height=\heightplot]{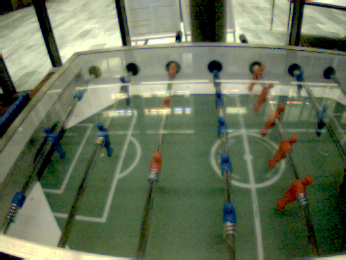}
    & \includegraphics[width=\widthplot,height=\heightplot]{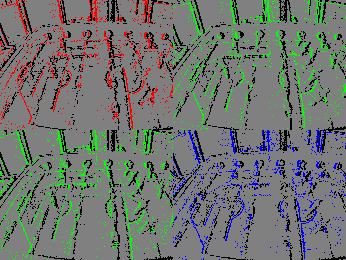}
    & & \rotatebox{90}{Country}
    & \includegraphics[width=\widthplot,height=\heightplot]{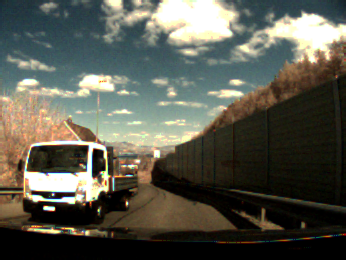}
    & \includegraphics[width=\widthplot,height=\heightplot]{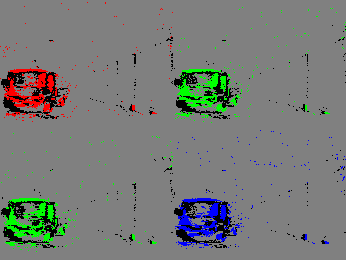}
    \\ 
   
   \rotatebox{90}{Corridor}
    & \includegraphics[width=\widthplot,height=\heightplot]{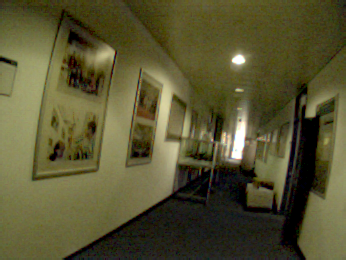}
    & \includegraphics[width=\widthplot,height=\heightplot]{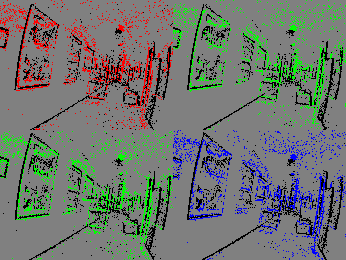}
   & & \rotatebox{90}{Tunnel}
    & \includegraphics[width=\widthplot,height=\heightplot]{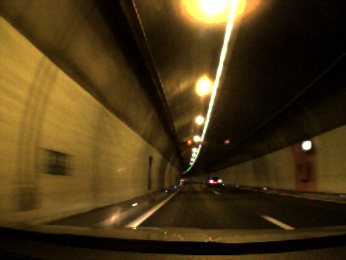}
    & \includegraphics[width=\widthplot,height=\heightplot]{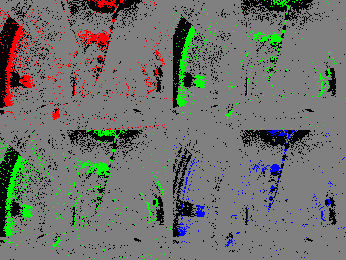}
   \\ 
       
    & DAVIS frame 
    & Events
    & & & DAVIS frame 
    & Events
    \\
    
    \end{tabular}
\caption{Impressions of the scenes from our dataset.
	Left column: color DAVIS frame; right column: color events.
}
\label{fig:dataset_previews}
\end{figure*}

\setlength{\tabcolsep}{0.2cm} %
\renewcommand{\arraystretch}{1}
\begin{table*}[t]
\centering
\caption{Types of scenes in our Color Event Camera Dataset.}
\label{tab:types}
\begin{tabular}{ p{1.3cm} C{0.4cm} C{0.4cm} p{1.3cm} p{7cm} p{3cm} }
	\theadCustom{{Type}}
	& \theadCustom{{\# Seq}}
	& \theadCustom{\makecell{{Length}\hspace{-2mm} \\ {(mins)}}}
	& \theadCustom{{Lux}}
	& \theadCustom{{Description}}
	& \theadCustom{{Possible Applications}}\\

	\midrule
	Simple
	& 16
	& 5
	& 80 - 1e3
	& Simple camera motions looking at simple objects and scenes with vibrant colors such as fruit, blocks and posters.
	& Image reconstruction \\
	\midrule
	Indoors
	& 15
	& 5
	& 0.8 - 1e3
	& Natural indoor scenes including office, kitchen, rooms and corridors.
	& Object detection  \\
	\midrule
	People 		
	& 27 	
	& 10
	& 400
	& Common actions and gestures such as sitting, waving, jumping, air guitar.
	& Action recognition  \\
	\midrule
	Driving 	
	& 12 	
	& 28 
	& 200 - 1e5
	& Footage from front windshield of car driving around country, suburban and city landscapes.
	Features tunnels, traffic lights, vehicles and pedestrians during the day in sunny conditions.
	& Segmentation, \linebreak Optical flow \\
	\midrule
	Calibration 		
	& 14	
	& 2	
	& 80 - 1e5
	& ColorChecker and density step target: indoors, outdoors, with and without infrared filter.
	& Color calibration  \\
	\midrule
	Simulated 	
	& - 	
	& -   	
	& - 
	& Color ESIM (adapted from \cite{Rebecq18corl}).
	Simulator can be used to generate unlimited sequences with ground truth depth, ego-motion, optical flow and more.
	& Optical flow, SLAM, Image reconstruction  \\
	\midrule
\end{tabular}
\end{table*}
\renewcommand{\arraystretch}{1}
\setlength{\tabcolsep}{0.05cm} %

\global\long\def\heightplot{3cm} %
\global\long\def\widthplot{4cm} %
\global\long\def\vTextWidth{0.4cm} %
\global\long\def\mColumnWidth{4.02cm} %
\begin{figure*}[t]
	\centering
	\begin{tabular}{  %
		>{\centering\arraybackslash}m{\vTextWidth}
		>{\centering\arraybackslash}m{\mColumnWidth}
		>{\centering\arraybackslash}m{\mColumnWidth}
		>{\centering\arraybackslash}m{\mColumnWidth}
		>{\centering\arraybackslash}m{\mColumnWidth}
		}
		
     \rotatebox{90}{Synthetic Room}
    & \includegraphics[width=\widthplot,height=\heightplot]{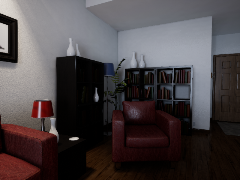}
    & \includegraphics[width=\widthplot,height=\heightplot]{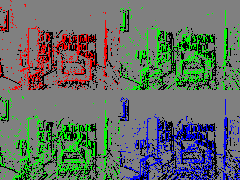}
    & \includegraphics[width=\widthplot,height=\heightplot]{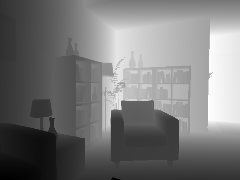}
    & \includegraphics[width=\widthplot,height=\heightplot]{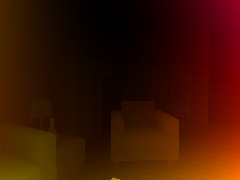}\\
    
    & (a) Frame & (b) Events & (c) Depth Map & (d) Optical Flow\\
    \end{tabular}
\caption{Example color events and ground truth modalities simulated with our color extension for ESIM. This scene was generated using the photorealistic rendering engine based on Unreal Engine.
}
\label{fig:esim_color}
\end{figure*}%

\textbf{Color Event Simulator.}
In addition to the real event datasets, we extended the event camera simulator ESIM \cite{Rebecq18corl} to allow simulation of color events%
\footnote{\url{https://github.com/uzh-rpg/rpg_esim}}.
Our extension operates on the ground-truth color (RGB) frames generated by the rendering engine, and simulates a color filter array (specifically, an RGBG Bayer pattern, as in the DAVIS346 used for this dataset).
The simulated Bayered frames are then processed by the event simulation code in ESIM, thus producing color events in the same way as the DAVIS346.
ESIM can readily provide multiple ground truth modalities, such as color frames, depth maps, optical flow maps, camera poses and camera velocities.
Our extension is compatible with all the rendering engines already bundled with ESIM, including a photorealistic rendering engine.
Figure~\ref{fig:esim_color} shows an example of color event data and ground truth modalities simulated by our extension of ESIM.
\section{Color Video Reconstruction}
\label{sec:reconstruction}

Image reconstruction from events serves two primary functions: 1) as a way to visualise events and 2) for use in downstream vision applications \eg object detection.

\subsection{Method}

\begin{figure*}
	\centering
	\begin{tabular}{  %
			>{\centering\arraybackslash}m{\vTextWidth}
			>{\centering\arraybackslash}m{\mColumnWidth}
			>{\centering\arraybackslash}m{\mColumnWidth}
			>{\centering\arraybackslash}m{\mColumnWidth}
			>{\centering\arraybackslash}m{\mColumnWidth}
		}
    
    \rotatebox{90}{\colorKeyboardLong}
    & \includegraphics[width=\widthplot,height=\heightplot]{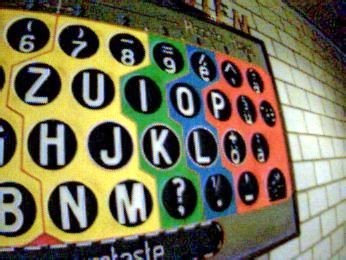}
    & \includegraphics[width=\widthplot,height=\heightplot]{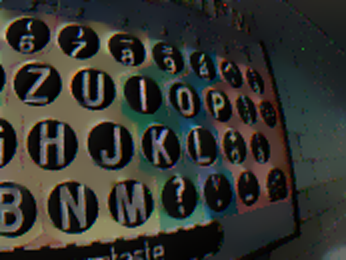}
    & \includegraphics[width=\widthplot,height=\heightplot]{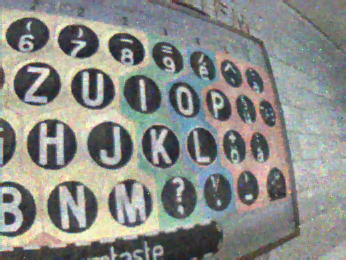}
    & \includegraphics[width=\widthplot,height=\heightplot]{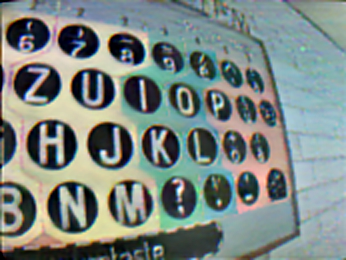}\\
    
	\rotatebox{90}{\jengaUnbroken}
	& \includegraphics[width=\widthplot,height=\heightplot]{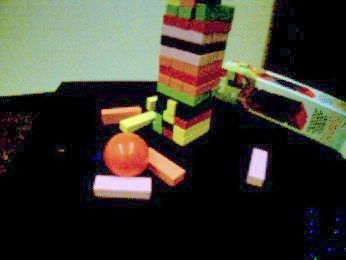}
	& \includegraphics[width=\widthplot,height=\heightplot]{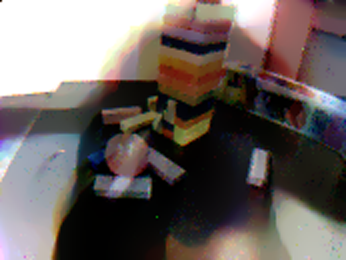}
	& \includegraphics[width=\widthplot,height=\heightplot]{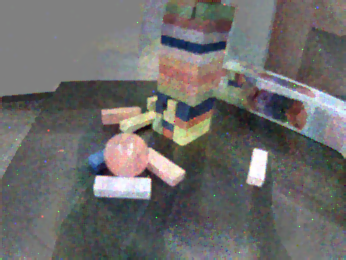}
	& \includegraphics[width=\widthplot,height=\heightplot]{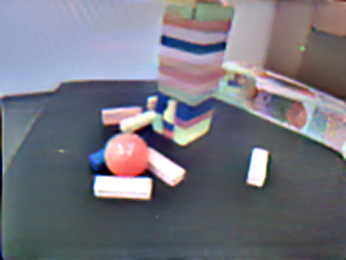}\\

    \rotatebox{90}{\flyingRoom}
    & \includegraphics[width=\widthplot,height=\heightplot]{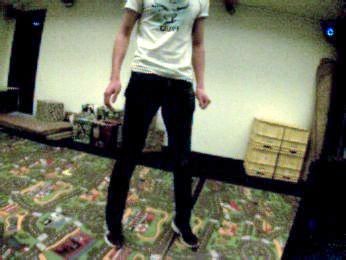}
    & \includegraphics[width=\widthplot,height=\heightplot]{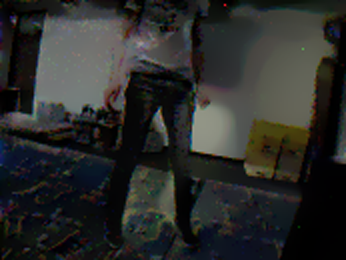}
    & \includegraphics[width=\widthplot,height=\heightplot]{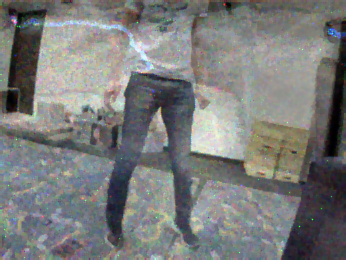}
    & \includegraphics[width=\widthplot,height=\heightplot]{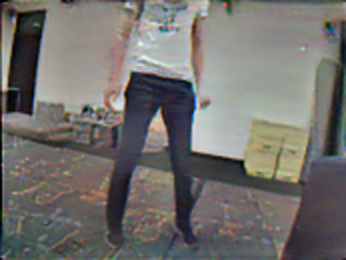}\\
    
    \rotatebox{90}{Driving}
    & \includegraphics[width=\widthplot,height=\heightplot]{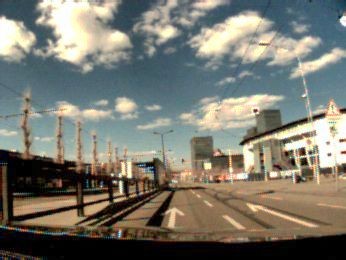}
    & \includegraphics[width=\widthplot,height=\heightplot]{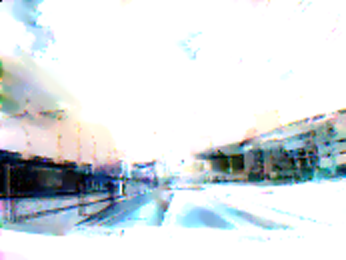}
    & \includegraphics[width=\widthplot,height=\heightplot]{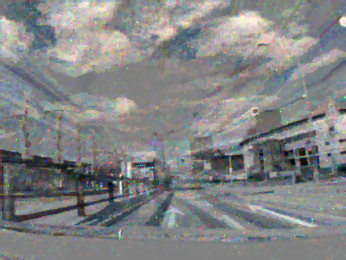}
    & \includegraphics[width=\widthplot,height=\heightplot]{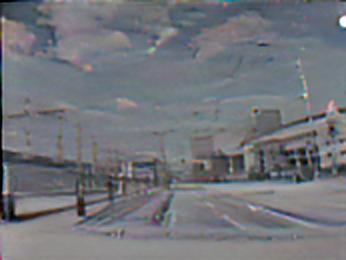}\\

    & \small (a) DAVIS frame & \small (b) MR \cite{Reinbacher16bmvc} & \small (c) HF \cite{Scheerlinck18accv} & \small (d) E2VID \cite{Rebecq19cvpr} \\
    \end{tabular}
\caption{Qualitative comparison of different color video reconstruction methods on our dataset (images randomly selected).
	Only events were used for each reconstruction method.
	Results (c), (d) qualitatively match the DAVIS frame (a).
}
\label{fig:reconstruction_results}
\end{figure*}

We evaluate and compare three state-of-the-art event-based image reconstruction methods on our Color Event Camera Dataset.
While these methods were originally designed for monochrome events, we found that with minimal modification all three were able to produce convincing color reconstructions.
While ``ground-truth'' color DAVIS frames were available, only color events were used as input to each method.

\textbf{1. Manifold Regularisation (MR).}%
\footnote{\url{https://github.com/VLOGroup/dvs-reconstruction}}
Reinbacher \etal \cite{Reinbacher16bmvc} use integration with spatio-temporal smoothing to recover image frames from events.
They use the surface of active events \cite{Benosman14tnnls} to define a manifold that guides regularisation.
We use default parameters provided by the authors; the integration window length is set to to $1,000$ events.

\textbf{2. High-pass Filter (HF).}%
\footnote{\url{https://github.com/cedric-scheerlinck/dvs_image_reconstruction}}
Scheerlinck \etal \cite{Scheerlinck18accv} show that a lightweight, asynchronous complementary filter can be used to obtain a continuous-time video from events and frames.
If desired, the frame input to the filter can be set to zero, resulting in a simple high-pass filter that produces reasonable results from only events.
Since each pixel is treated independently without spatial smoothing, the Bayer pattern is preserved, and demosaicing \cite{Kimmel99tip} can be used to recover an RGB image at any point in time.
We use a gain of 0.06 for both \texttt{cutoff\_frequency} and \texttt{cutoff\_frequency\_per\_event\_component}.
As a final post-processing step, we apply a 5$\times$5 bilateral filter with \texttt{spatial\_filter\_sigma} set to 1.0 for each output reconstruction.

\textbf{3. E2VID Neural Network (E2VID).}
Rebecq \etal \cite{Rebecq19cvpr} show that a recurrent neural network trained on a large amount of event data simulated with ESIM \cite{Rebecq18corl} can generate high quality video reconstructions from event data only.
E2VID converts the stream of events into a sequence of ``event tensors'', each consisting of a fixed batch of events represented as a 3D spatio-temporal voxel grid.
The sequence of event tensors is passed to a recurrent UNet that outputs a sequence of reconstructed image frames.

Manifold regularization (MR) and E2VID utilize spatial smoothing, which destroys the Bayer pattern if applied directly to events.
For both of these methods, we found that color images can still be obtained by reconstructing red, green and blue channels independently (at quarter resolution), then upsampling to the original resolution using bicubic interpolation.
Because of the Bayer pattern, the four different (upsampled) color channels will not be exactly aligned.
Therefore, we shift each color channel by one pixel horizontally and/or vertically so that all four color channels are geometrically aligned.
We fuse both green channels (after alignment) by simply taking the mean.
In contrast, the High-pass filter (HF) treats each pixel independently and does not perform spatial smoothing.
Thus, it can be applied directly to events, then converted to color using demosaicing \cite{Kimmel99tip}.
\subsection{Results}
\label{sec:results}

Figure \ref{fig:reconstruction_results} shows reconstruction results of all three methods; Manifold regularisation (MR), High-pass filter (HF) and events-to-video neural network (E2VID), alongside DAVIS frames from the Color-DAVIS346.
HF and E2VID preserve color well and qualitatively match the DAVIS frame.
We encourage the reader to watch the accompanying video, which convey our results better than still-images.

\begin{figure*}[t]
	\centering
	\begin{tabular}{  %
			>{\centering\arraybackslash}m{\vTextWidth}
			>{\centering\arraybackslash}m{\mColumnWidth}
			>{\centering\arraybackslash}m{\mColumnWidth}
			>{\centering\arraybackslash}m{\mColumnWidth}
			>{\centering\arraybackslash}m{\mColumnWidth}
		}
    \rotatebox{90}{Initialization}
    & \includegraphics[width=\widthplot,height=\heightplot]{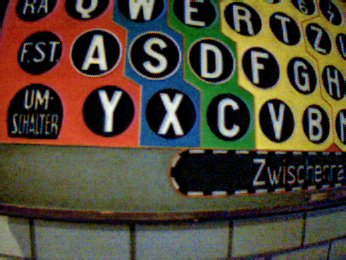}
    & \includegraphics[width=\widthplot,height=\heightplot]{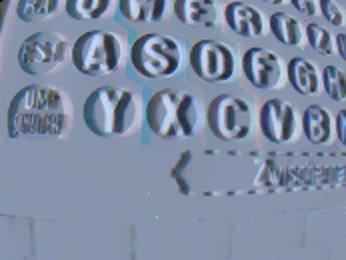}
    & \includegraphics[width=\widthplot,height=\heightplot]{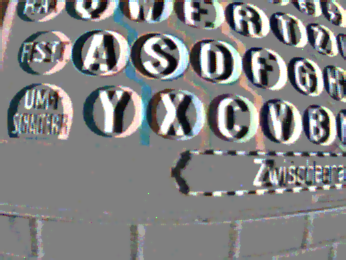}
    & \includegraphics[width=\widthplot,height=\heightplot]{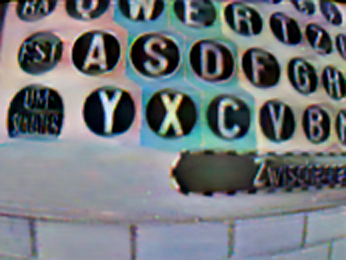}\\
    
	\rotatebox{90}{Fast motion}
    & \includegraphics[width=\widthplot,height=\heightplot]{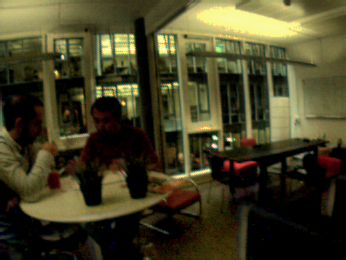}
    & \includegraphics[width=\widthplot,height=\heightplot]{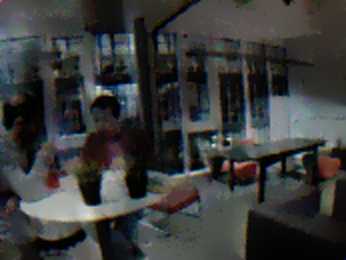}
    & \includegraphics[width=\widthplot,height=\heightplot]{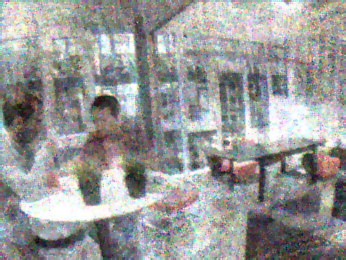}
    & \includegraphics[width=\widthplot,height=\heightplot]{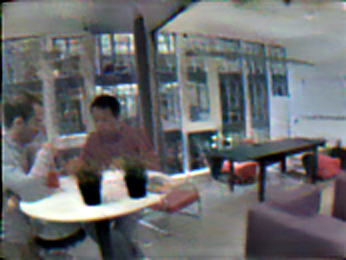}\\
	
	\rotatebox{90}{Sharpness}
    & \includegraphics[width=\widthplot,height=\heightplot]{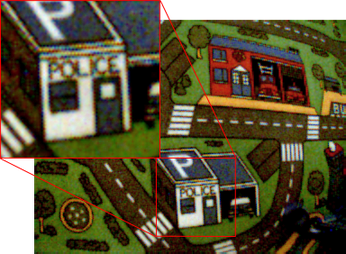}
    & \includegraphics[width=\widthplot,height=\heightplot]{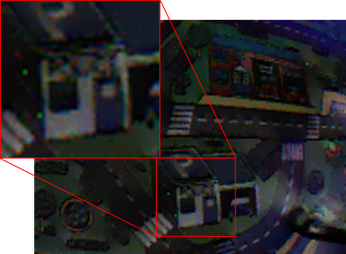}
    & \includegraphics[width=\widthplot,height=\heightplot]{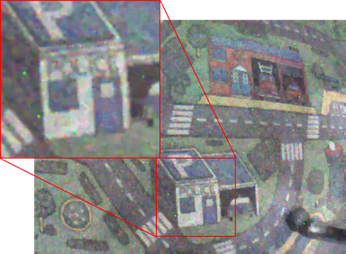}
    & \includegraphics[width=\widthplot,height=\heightplot]{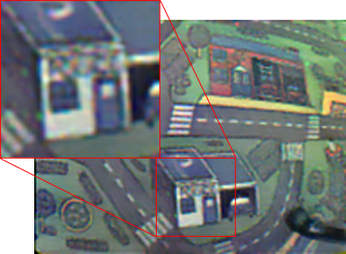}\\
    
	\rotatebox{90}{Memory}
    & \includegraphics[width=\widthplot,height=\heightplot]{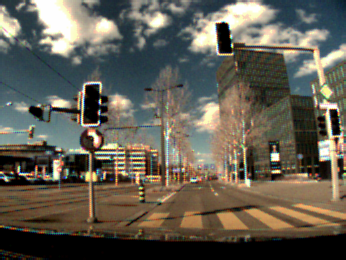}
    & \includegraphics[width=\widthplot,height=\heightplot]{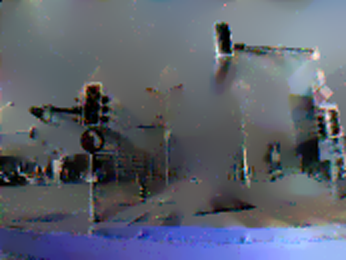}
    & \includegraphics[width=\widthplot,height=\heightplot]{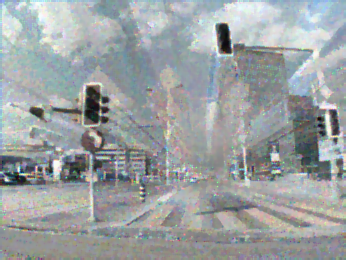}
    & \includegraphics[width=\widthplot,height=\heightplot]{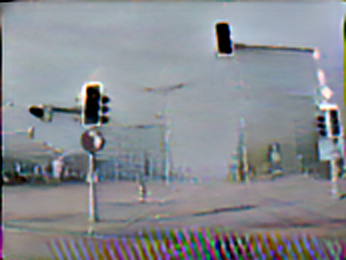}\\    

	\rotatebox{90}{HDR}
    & \includegraphics[width=\widthplot,height=\heightplot]{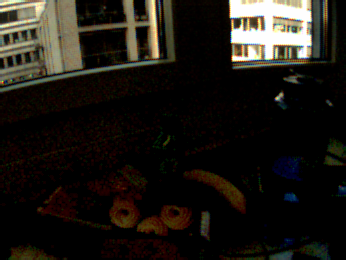}
    & \includegraphics[width=\widthplot,height=\heightplot]{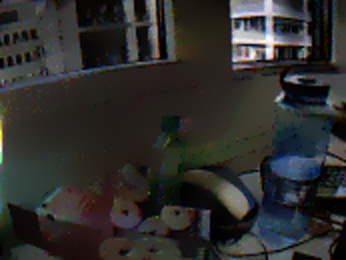}
    & \includegraphics[width=\widthplot,height=\heightplot]{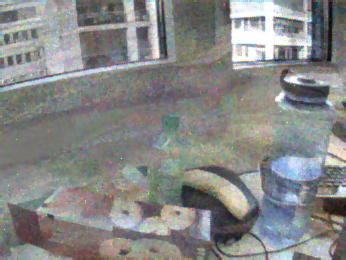}
    & \includegraphics[width=\widthplot,height=\heightplot]{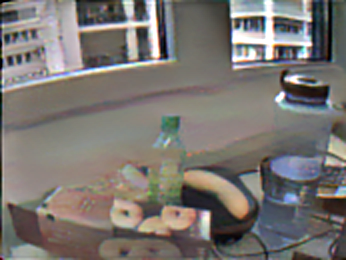}\\

	\rotatebox{90}{Low light}
    & \includegraphics[width=\widthplot,height=\heightplot]{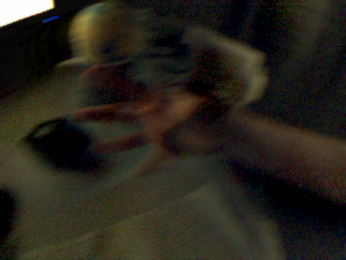}
    & \includegraphics[width=\widthplot,height=\heightplot]{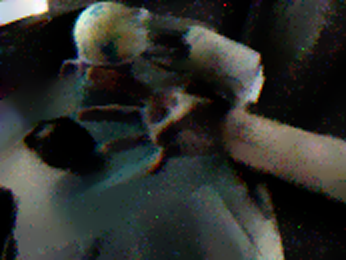}
    & \includegraphics[width=\widthplot,height=\heightplot]{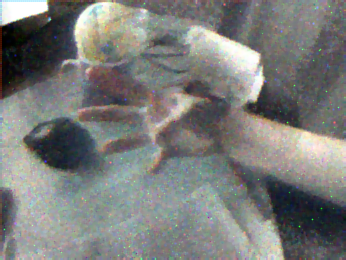}
    & \includegraphics[width=\widthplot,height=\heightplot]{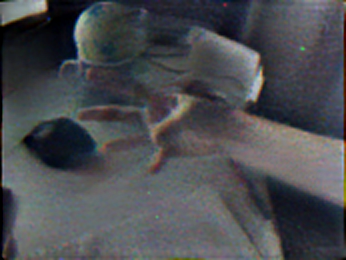}\\
    
    & \small (a) DAVIS frame & \small (b) MR \cite{Reinbacher16bmvc} & \small (c) HF \cite{Scheerlinck18accv} & \small (d) E2VID \cite{Rebecq19cvpr}\\
    \end{tabular}
\caption{Edge cases for different reconstruction methods.
	First row: initialization, all method but E2VID fail.
	Second row: fast motion, HF accumulates more noise.
	Third row: zoom on carpet, HF preserves fine details better.
	Fourth row: Low apparent motion \eg in the sky, HF preserves slow moving objects better.
	Fifth row: HDR scene, DAVIS cannot capture entire intensity range, reconstructions can.
	Sixth row: dark room (2 lux), DAVIS suffers motion blur, not the reconstructions.
}
\label{fig:reconstruction_results_edge_cases}
\end{figure*}

Figure \ref{fig:reconstruction_results_edge_cases} displays edge cases such as high-speed, HDR \etc that highlight strengths and weaknesses of each reconstruction method and the DAVIS frames:

\textbf{Initialisation (first row).}
Both MR and HF are initialised at zero and rely on integration of events to build a consistent image over time. Thus, they are prone to producing edge-like images, particularly within the first few milliseconds after initialisation, until enough events `fill in' the missing information.
In contrast, E2VID is good at filling in gaps and can hallucinate color accurately in places with no events.

\textbf{Fast Motion (second row).}
HF is a temporal high-pass filter, and is sensitive to temporal components in the input signal, such as frequency and speed.
Thus, the quality of the reconstruction can be adversely affected by extremely fast (or slow) motions.%

\afterpage{\clearpage}
\clearpage
In addition, fast motions tend to generate noise in the event stream that is accumulated without discrimination by the integrator in HF.
MR and E2VID are good at rejecting noise from fast motion and showcase the attractive properties of event cameras for challenging scenarios.

\textbf{Sharpness (third row).}
MR and E2VID rely on spatial smoothing to filter out noise from the event stream, which can degrade sharpness of fine details.
For color reconstruction, the spatial smoothing property of these two methods destroys the Bayer pattern, requiring each color to be reconstructed independently (at quarter resolution), then upscaled back to the original resolution, further losing fine details.
In contrast, HF requires no spatial smoothing, so a raw intensity reconstruction at full resolution is possible, since the Bayer pattern is preserved.
A demosaicing algorithm \cite{Kimmel99tip} can be used to convert the raw output to color without loss of resolution, resulting in a sharper reconstruction.

\textbf{Memory (fourth row).}
The ``memory'' (\ie the time span over which information in the event data can be propagated) is variable between all three methods.
For HF, the size of the temporal receptive field (memory) is explicitly encoded through the cutoff frequency parameter.
Hence, the duration across which information can be propagated can be set to an arbitrarily high amount of time, at the expense of integrating more noise, and creating ``bleeding'' patterns following moving objects.
By contrast, MR and E2VID have an implicit memory, whose size can vary with the number of events used in each integration window (MR), or event tensor (E2VID).
However, we observe that the memory of MR and HF is notably smaller than HF, which is particularly visible in the driving sequence (fourth row of Fig.~\ref{fig:reconstruction_results_edge_cases}), where HF is able to reconstruct slow moving objects, \eg the clouds or the distant buildings, in contrast to MR and E2VID.

\textbf{HDR (fifth row).}
Since the APS is limited to a uniform exposure duration for all pixels, the DAVIS frame has low dynamic range compared to events.
Thus, dark regions are often underexposed while bright regions (window) are well exposed, and vice versa.
Reconstructions from MR, HF and E2VID all showcase the high dynamic range property of events, \ie both dark and bright regions are clear.

\textbf{Low light (sixth row).}
Low lighting is a challenge for conventional cameras because the exposure duration must be increased to avoid underexposure, leading to motion blur.
While the DAVIS frame is motion blurred, MR, HF and E2VID demonstrate immunity to motion blur, even in challenging low lighting conditions.

\subsection{Application of Reconstructions}
While many computer vision algorithms work on grayscale images, it is well established that incorporating color information can significantly boost performance for the task at hand \cite{Gevers12book}.
This is because color images contain more information about the scene than grayscale images, which can only encode structural information.
This is particularly true in recognition tasks, where color can be an important visual cue.
Figure~\ref{fig:yolo} shows one example where color improves object detection performance.
We apply YOLO \cite{Redmon16cvpr} to E2VID images reconstructed from both grayscale and color events and observe that color offers qualitative improvement.
While image reconstructions can be used directly for the task at hand, they may also be used to generate proxy labels (\eg segmentation, optical flow, recognition) that can be transferred to events.

\global\long\def\frontWidth{4.05cm} %
\begin{figure}[t]
	\centering
    \begin{tabular}{cc}

    \includegraphics[width=\frontWidth]{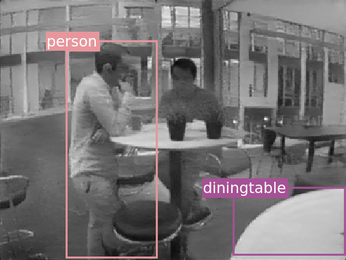}
    & \includegraphics[width=\frontWidth]{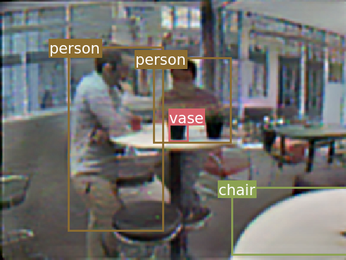}
    \\
 
    \end{tabular}
    \caption{Object detection (YOLO \cite{Redmon16cvpr}) on reconstructed images using E2VID.
    	Color (right) tends to improve detection performance.
    }
    \label{fig:yolo}
\end{figure}%

\section{Conclusion}
\label{sec:conclusion}
\vspace{-2mm}
We present the first \emph{Color Event Camera Dataset}, containing both frames and events across a diverse range of scenes, motions and lighting conditions.
We release an open source color event camera simulator based on ESIM \cite{Rebecq18corl}.
We show how three state-of-the-art event-based image reconstruction methods can be adapted for color video reconstruction, and compare strengths/weaknesses of each method.
We hope that our Color Event Camera Dataset and simulator will inspire future work with color events, which we believe is the next step for event-based vision.
\section*{Acknowledgements}
We would like to thank Prof. Tobi Delbruck and the Sensors group at the Institute of Neuroinformatics (ETH \& University of Zurich), and Inivation for providing the camera.
This work was supported by
(i) the Australian Government Research Training Program Scholarship 
(ii) the Australian Research Council through the ``Australian Centre of Excellence for Robotic Vision'' under Grant CE140100016
(iii) the Swiss Government Excellence Scholarship
(iv) the Swiss National Center of Competence Research Robotics (NCCR) 
(v) Qualcomm (through the Qualcomm Innovation Fellowship Award 2018)
(vi) the SNSF-ERC Starting Grant.

\clearpage
{\small
\balance
\bibliographystyle{ieeetr} %
\bibliography{all,extra_refs}
}

\end{document}